# Remote Sensing Cross-Modal Text-Image Retrieval Based on Global and Local Information

Zhiqiang Yuan⬝, *Student Member, IEEE*, Wenkai Zhang⬝, *Member, IEEE*, Changyuan Tian, Xuee Rong⬝,
Zhengyuan Zhang⬝, Hongqi Wang, *Member, IEEE*, Kun Fu, *Member, IEEE*,
and Xian Sun⬝, *Senior Member, IEEE*

*Abstract*—**Cross-modal remote sensing text-image retrieval (RSCTIR) has recently become an urgent research hotspot due to its ability of enabling fast and flexible information extraction on remote sensing (RS) images. However, current RSCTIR methods mainly focus on global features of RS images, which leads to the neglect of local features that reflect target relationships and saliency. In this article, we first propose a novel RSCTIR framework based on global and local information (GaLR), and design a multi-level information dynamic fusion (MIDF) module to efficaciously integrate features of different levels. MIDF leverages local information to correct global information, utilizes global information to supplement local information, and uses the dynamic addition of the two to generate prominent visual representation. To alleviate the pressure of the redundant targets on the graph convolution network (GCN) and to improve the model's attention on salient instances during modeling local features, the denoised representation matrix and the enhanced adjacency matrix (DREA) are devised to assist GCN in producing superior local representations. DREA not only filters out redundant features with high similarity, but also derives more powerful local features by enhancing the features of prominent objects. Finally, to make full use of the information in the similarity matrix during inference, we come up with a plug-and-play multivariate rerank (MR) algorithm. The algorithm utilizes the $k$ nearest neighbors of the retrieval results to perform a reverse search, and improves the performance by combining multiple components of bidirectional retrieval. Extensive experiments on public datasets strongly demonstrate the state-of-the-art performance of GaLR methods on the RSCTIR task. The code of GaLR method, MR algorithm, and corresponding files have been made available at: https://github.com/xiaoyuan1996/GaLR.**

Manuscript received December 28, 2021; revised March 4, 2022; accepted March 27, 2022. Date of publication March 31, 2022; date of current version April 15, 2022. This work was supported by the National Science Fund for Distinguished Young Scholars under Grant 67125105. *(Corresponding author: Xian Sun.)*

Zhiqiang Yuan, Changyuan Tian, Xuee Rong, and Zhengyuan Zhang are with the Aerospace Information Research Institute, Chinese Academy of Sciences, Beijing 100190, China, also with the Key Laboratory of Network Information System Technology (NIST), Institute of Electronics, Chinese Academy of Sciences, Beijing 100190, China, also with the University of Chinese Academy of Sciences, Beijing 100190, China, and also with the School of Electronic, Electrical and Communication Engineering, University of Chinese Academy of Sciences, Beijing 100190, China (e-mail: yuanzhiqiang19@mails.ucas.ac.cn; tianchangyuan21@mails.ucas.ac.cn; rongxuee19@mails.ucas.ac.cn; zhangzhengyuan16@mails.ucas.ac.cn).

Wenkai Zhang, Hongqi Wang, Kun Fu, and Xian Sun are with the Aerospace Information Research Institute, Chinese Academy of Sciences, Beijing 100190, China, and also with the Key Laboratory of Network Information System Technology (NIST), Institute of Electronics, Chinese Academy of Sciences, Beijing 100190, China (e-mail: zhangwk@aircas.ac.cn; wiecas@sina.com; kunfuiecas@gmail.com; sunxian@mail.ie.ac.cn).

Digital Object Identifier 10.1109/TGRS.2022.3163706

## I. INTRODUCTION

IN RECENT years, researchers pay more and more attention to remote sensing (RS) technology, which has been widely used in the fields of resource acquisition, military reconnaissance, and disaster monitoring [1], [2]. Meanwhile, the rapid developments of RS applications have led to an explosive growth of RS images [3], [4]. To extract relevant images from the exponential RS data using different modal information, effective automatic RS cross-modal retrieval (RSCR) methods are increasingly needed [5]–[7]. Due to the utility of text in human-computer interaction, RS cross-modal text-image retrieval (RSCTIR) has received unprecedented attention automatically [8], [9].

Current RSCTIR methods are mainly divided into caption-based methods and embedded-based methods [10]. Caption-based RSCTIR tends to use a caption generator to generate RS labels, and obtains the retrieval results by calculating the bilingual evaluation understudy (BLEU) [11] score between the query text and the generated labels. To make the generated caption interpretable, Wang *et al.* [12] designed an explainable word–sentence framework, decomposing the task into the word classification and sorting task. To get the caption generator with a more comprehensive semantic understanding of complex RS images, Li *et al.* [13] proposed recurrent attention and semantic gate framework to generate better context vector. Different from the two-stage caption-based method, embedded-based RSCTIR directly projects the RS image and the query text into the same high-dimensional space to calculate the feature distance. Due to the heterogeneity gap, the construction of a well-represented unimodal embedding matrix in this method has become an urgent problem for researchers. Yuan *et al.* [14] devised a multi-scale visual self-attention module to filter the salient features in redundant RS images and utilized a cross-modal guidance mechanism to get a superior multi-modal representation. To fully excavate the latent correspondence between RS images and text, Cheng *et al.* [9] proposed a semantic alignment module to filter and optimize data features to achieve higher retrieval accuracy. The embedded-based RSCTIR calculates the cross-modal similarity in a single stage, which greatly reduces the loss of information transformation. Also due to the widespread use of the embedded-based approaches in semantic localization tasks [15], this method has recently become the preferred retrieval mode in RSCR tasks.

Although embedded-based RSCTIR methods have achieved acceptable accuracy, there are still some problems that need to be overcome. Firstly, current RS retrieval frameworks are mainly based on global features of images [16], [17], while local features reflecting the relationships between objects are often ignored. The use of global features gives the model a better understanding of the overall image, but leaves it lacking in fine-grained awareness of the objects. In contrast, the utilization of local information constructed by targets will enable the model to have a good perception of fine-grained objects and corresponding relationships [18]. These two types of features represent RS images from high and low levels, respectively, and using the feature at a single level will inevitably lead to a lack of information. To solve this problem, we propose a RSCTIR framework based on global and local features, and design a multi-level information dynamic fusion (MIDF) module to marriage information from different levels. MIDF utilizes local information to supplement global information, uses global information to correct local information, and generates prominent visual representation through dynamic addition.

Secondly, unlike natural scenes, RS images with high redundancy will inevitably result in information redundancy if all object information is leveraged directly for relationship modeling [19], [20]. On the one hand, too many objects reduce the efficiency of the model, while on the other hand, the model cannot focus on the salient instance information in the RS image. To reduce the modeling pressure caused by redundant targets to graph convolution networks (GCNs) and increase the model's attention to salient instances, a denoised representation matrix and a enhanced adjacency matrix (DREA) are devised to assist GCN in producing good local representations. DREA filters out redundant features with high similarity and enhances the features of salient targets, which can obtain more powerful local features and thus improving the retrieval efficiency.

Thirdly, the previous retrieval frameworks ignore the bidirectional ranking information in the cross-modal similarity matrix during inference [21], which could be utilized for secondary optimization of the retrieval results. The retrieval process is reversible, if sample $A$ in modal $\mathbb{A}$ is a retrieval candidate for sample $B$ in modal $\mathbb{B}$, the reverse is also true. Based on this, we propose a multivariate rerank (MR) algorithm, which uses $k$ candidates of another modality for reverse retrieval, and combines multiple factors of bidirectional retrieval to optimize results. The suggested post-processing algorithm comprehensively integrates the ranking information of $i2t$ and $t2i$ in the original similarity matrix, while adding a significant component to further enhance the reranking accuracy.

The main contributions of our work are as follows.

- To equip the model with the capability of a multi-level and comprehensive understanding of RS images, a RSCTIR model based on global and local information (GaLR) is established. At the same time, a novel MIDF module for dynamic fusion of multi-level information is designed, which leverages local information to correct global information, utilizes global information to supplement local information, and uses the dynamic addition of the two to generate prominent visual representation.

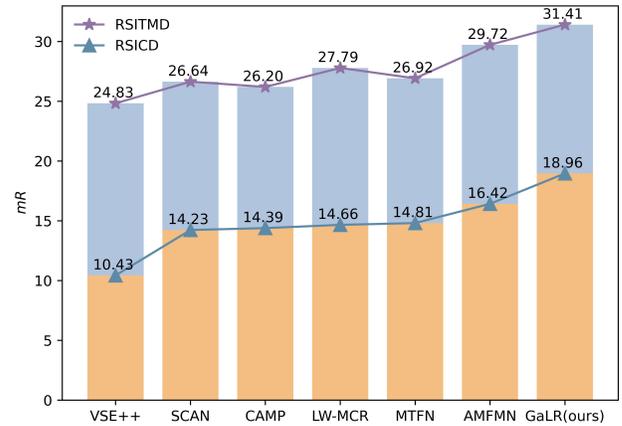

Fig. 1. Comparison of the proposed GaLR with other methods on two popular RS image-text datasets. Our method achieves the accuracy of state-of-the-art on the RSICD and RSITMD datasets, which greatly improves the RS cross-model retrieval performance.

- To alleviate the pressure on the model from redundant target relations and increase the model's focus on salient instances, we come up with DREA to assist the GCN in producing better local representations. DREA filters the redundant features with high similarity and enhances the features of salient targets, which enables GaLR to obtain more transcendent visual representation.

- To make full use of the information in the similarity matrix, a plug-and-play MR algorithm is devised to make the retrieval results have a secondary optimization. The post-processing algorithm comprehensively considers the various information in the similarity matrix through a bidirectional retrieval method, and can make the model obtain higher retrieval accuracy without extra training.

We have performed plenty of comparative experiments on several RS text-image datasets to demonstrate the state-of-the-art performance of the proposed GaLR as shown in Fig. 1. In addition, multiple ablation experiments are conducted to analyze the sakes for the superior performance of DREA and MR. Subsequently, we first present related work on RS text-image retrieval and graph neural network (GNN) in Section II. Then, Section III introduces a detailed description of the suggested models and algorithms. Further, we conduct a large number of experiments to verify the effectiveness of the proposed method in Section IV. Finally, the conclusions are given in Section V.

## II. RELATED WORK

This section first introduces the development of RSCTIR, and then reviews some methods of GNN in relation modeling.

### A. RS Cross-Modal Text-Image Retrieval (RSCTIR)

RSCTIR refers to recalling required RS images with text. The semantic gap caused by the heterogeneity of multimodal data makes the RSCTIR tasks extremely challenging. In terms of implementation methods, RSCTIR tasks can be generally divided into caption-based methods and embedded-based methods.

Caption-based methods can be regarded as two-stage retrieval methods. These methods usually first generates annotations for each RS image in the database by the caption generator, and then calculates the similarity between the query text and the generated annotations using BLEU [11] in the retrieval phase. Qu *et al.* [4] firstly utilized multi-modal deep networks for semantic understanding of the high resolution RS images. To better characterize RS images, Shi and Zou [22] proposed an RS image captioning (RSIC) framework based on deep learning and fully convolution network to semantically decompose ground elements at different scales. Lu *et al.* [23] contributed a large RSIC dataset and conducted a comprehensive review to fully advance the task of RSIC. Sumbul *et al.* [24] proposed a summary driven RSIC method to overcome information deficiencies and validated the effect on multiple RS text-image datasets. Li *et al.* [25] suggested a new truncated cross-entropy loss to alleviate the overfitting problem in RSIC. To solve the problem that the current caption generator requires high computing power, Genc *et al.* proposed a decoder based on support vector machine [26], which is effective when just a limited amount of training samples is available. Despite the relative maturity of caption-based RSCTIR [12], [27]–[29], the model inevitably introduces noise due to the limitations of dual-stage process, thus affecting the accuracy of retrieval.

The embedded-based RSCTIR methods refer to mapping RS image and text into the same high-dimensional space and measuring the cross-modal similarity by appropriate distance. To obtain a more robust embedding, Abdullah *et al.* [8] proposed a deep bidirectional triplet network to learn joint encoding between multiple modalities and utilized an averaging fusion strategy to fuse the features of multiple text-image pairs. To solve the problem of multi-scale scarcity and target redundancy in RSCITR, Yuan *et al.* [14] proposed an asymmetric multimodal feature matching network (AMFMN) and contributed a fine-grained RS image-text dataset for this task. By exploring the potential correspondence between RS images and text, Cheng *et al.* [9] proposed a semantic alignment module to get a more discriminative feature representation. Lv *et al.* [30] suggested a fusion-based correlation learning model for RS image-text retrieval, and solve the heterogeneity gap problem by knowledge distillation. From another perspective, Yuan *et al.* [15] proposed a lightweight image-text retrieval model to achieve faster RS cross-modal retrieval, and use the method of knowledge distillation and contrast learning to enhance the retrieval performance. Although some achievements have been made in this field [16], [31], the previous approaches tend to ignore the relationship among objects when performing visual representations, which leads to the models unable to fully exploit the semantics in RS image.

### B. Graph Neural Network (GNN)

More recently, there has been a surge of interest in GNN which can better represent relationships at the object level. To process the data in graph domain, Scarselli *et al.* [32] first proposed GNN model and the supervised learning algorithm were derived to estimate the parameters of the model. Gao *et al.* [33] put forward the learnable graph convolutional layer which improved the performance of regular convolutional operations on generic graphs. Pei *et al.* [34] introduced a geometric aggregation scheme for GNN which could retain the structural information of nodes in neighborhoods better and enhance the ability to capture long-range dependencies in disassortative graphs. Meanwhile, many of the current scholars on RS pay particular attention to GNN model. Based on the GCN, Wan *et al.* [35] utilized multi-scale dynamic GCN model for hyperspectral image classification. Similarly, Min *et al.* [36] established a nonlocal GCN which could offer competitive results and high-quality classification maps. Hong *et al.* [37] fully investigated the different performance of convolution neural network (CNN) and GCN in hyper-spectral image classification from both qualitative and quantitative perspectives. Liu *et al.* [38] designed the context graph attention network for ground-based RS cloud classification. For multi-label RS scene categorization, Kang *et al.* [39] proposed a new graph relation network (GRN), which improved the effectiveness in terms of classification and retrieval. GNN plays an increasingly important role in object-level relationship modeling [40].

## III. METHOD

This section introduces the proposed RSCTIR model based on GaLR. Fig. 2 shows the overall architecture of the GaLR. We present our method from four aspects: formulation, optimize local representation by DREA mechanism, MIDF, and MR. Then we have introduced the objective function and algorithm process.

### A. Formulation

When cross-modal retrieval is performed, the model needs to calculate the similarity between the RS images in the database and the query text. In the traditional approach, the image $I$ and text $T$ are usually encoded as two independent vectors which are mapped into the same space. This process can be represented as

$$v = W_v \ I, \quad t = W_t \ T \tag{1}$$

$$S = \cos(v, t) \tag{2}$$

where $W_v$ and $W_t$ are the weight matrix for the representation of visual and text. $v$ and $t$ denote the representation features of the visual and text separately. $\cos(x, y)$ denotes the cosine similarity of vector $x$ and vector $y$. $S$ is the cross-modal similarity. Next, we will present our method from visual representation and text representation.

*1) Visual Representation:* In the traditional method, CNNs are often used to construct the $W_v$ so as to embed the images [41]. Although these methods directly map the global information into the high-dimensional space, they ignore the distinction between significant objects and redundant information due to the complexity of RS image [14]. In addition, the relationship between objects in RS images can not be well represented by global features, which may lead to degradation of retrieval performance. In response to these problems, we add local feature branch to the traditional method to reflect the relationship between important objects in the RS image.

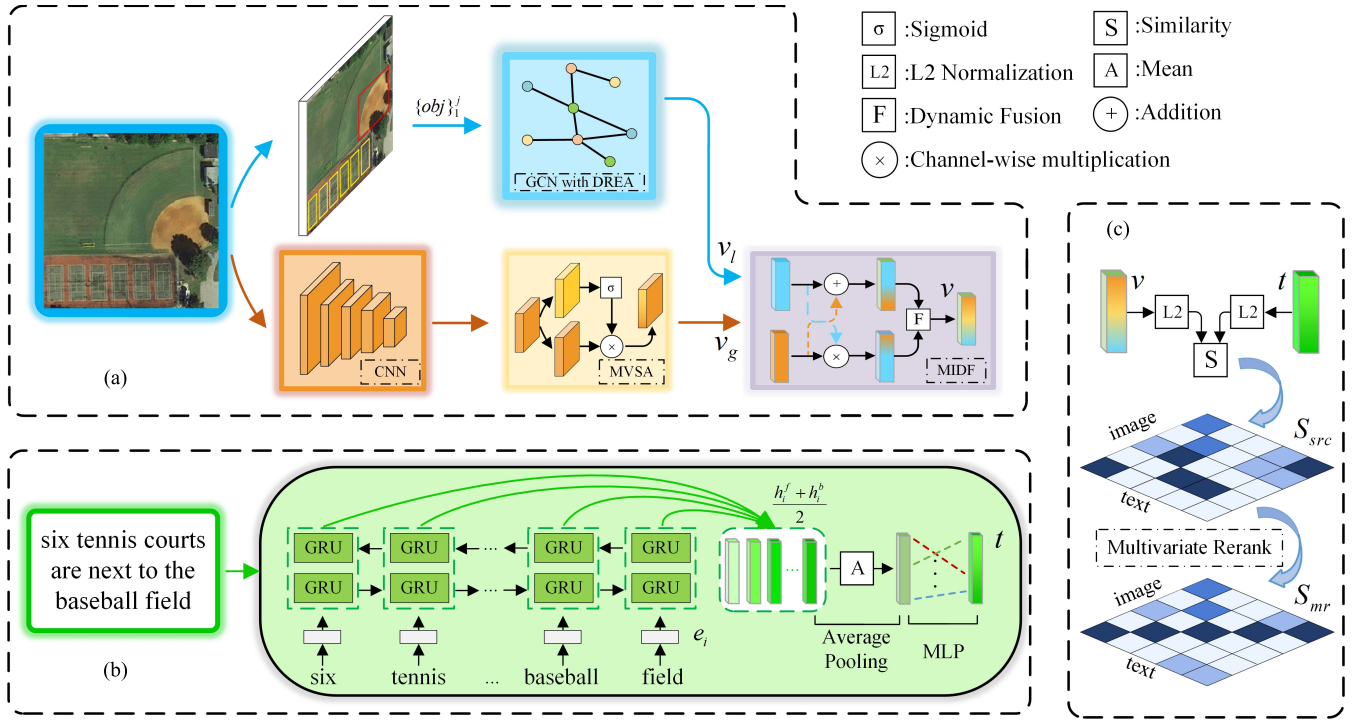

Fig. 2. Proposed RSCTIR framework based on GaLR. Compared with the retrieval models constructed with only global features, GaLR incorporates optimized local features in the visual encoding considering the target redundancy of RS. The multi-level information dynamic fusion module is designed to fuse the two types of information, using the global information to supplement the local information and utilizing the latter to correct the former. The suggested MR algorithm as a post-processing method further improves the retrieval accuracy without extra training. (a) Visual representation. (b) Text representation. (c) Post-processing.

Super RS image global features needs to pay attention to the salient object information and suppress useless feature expression. To obtain such global features, the multi-level visual self-attention (MVSA) proposed by Yuan *et al.* [14] came into being. For this purpose, we utilize convolution transformation and MVSA mask to extract global features

$$v_g = \mathcal{C}_{\text{mask}}(I) \qquad (3)$$

where $v_g$ is the global feature of the RS image, and $\mathcal{C}_{\text{mask}}$ is the CNN with MVSA. Then, to construct the relationship among the local features, we use the object detector pre-trained on the DOTA [42] dataset to extract the objects in the RS image

$$\{\text{obj}\}_1^j = \text{Detect}(I) \qquad (4)$$

where $\text{Detect}(x)$ represents extract the targets of the RS image by the object detection network, and $\{\text{obj}\}_1^j$ represents the targets extracted by the detector. Further, the GCN $\mathcal{G}$ is used to model the relationship among the targets, which can be represented as

$$v_l = \mathcal{G}\Big(\{\text{obj}\}_1^j\Big) \qquad (5)$$

where $v_l$ is the local feature of the RS image.

Global features tend to reflect the overall information of the image, while local features reflect the content in a more fine-grained manner at the target level. However, when there are few objects in the RS image, the local features obtained by the detector are weaker relative to the global features. While for RS images with no targets, local features often contain very little information, so global features are extremely important at this time. To achieve the dynamic fusion of global and local features, we propose a MIDF module

$$v = \text{MIDF}\big(v_g, v_l\big) \qquad (6)$$

MIDF leverages local information to correct global information, utilizes global information to supplement local information, and uses the dynamic addition of the two to generate prominent visual representation. The suggested fusion module comprehensively considers the difference between global and local features, and generates dynamic weights to weight these two features.

*2) Text Representation:* When modeling text, recurrent neural networks are utilized to extract temporal information. The query text $T$ included $n$ words is expressed as $\{w_i\}_1^n$. The words in the text $T$ are embedded into the word vector and fed into the RNN [43], and then the representation of the text $t$ is obtained through the multilayer perceptron MLP. The above process can be modeled as follows:

$$e_i = W_e(w_i)(i \in [1, n]) \qquad (7)$$

$$h_i^f = \text{GRU}^f\big(e_i, h_{i-1}^f\big) \qquad (8)$$

$$h_i^b = \text{GRU}^b\big(e_{n-i}, h_{i-1}^b\big) \qquad (9)$$

$$t = \text{MLP}\left(\frac{1}{n}\sum_{i=1}^{n}\frac{h_i^f + h_i^b}{2}\right) \quad (10)$$

where $W_e$ denotes the word embedding matrix, $e_i$ denotes the $i$th embedded word. $\text{GRU}^f(x)$ and $\text{GRU}^b(x)$ denotes the forward and the backward GRU [44]. $h_i^f$ and $h_i^b$ indicate the hidden state of the $\text{GRU}^f$ and $\text{GRU}^b$ at step $i$, separately.

### B. Optimize Local Representation by DREA Mechanism

Local features characterize RS images in a more fine-grained way. In this subsection, we use GCN to construct local representations after obtaining the objects in RS image, and propose a DREA mechanism to optimize extracted local features.

After using the object detector to extract instances of the RS image, to reflect the relationship between the extracted objects, a GCN [45] is leveraged to generate local features. Compare to traditional neural network, GCN uses the information of the node as the input of the network, and utilizes adjacency matrix to characterize the relationship between nodes. For the output $X^{(l+1)}$ of the $l$th GCN layers, we define

$$X^{(l+1)} = \sigma\left(\widetilde{D}^{-\frac{1}{2}}\widetilde{A}\,\widetilde{D}^{-\frac{1}{2}}X^{(l)}W^{(l)}\right) \quad (11)$$

where $W^{(l)}$ is the learnable weight matrix of the ($l$)th layers, and $\sigma$ is the activation function. The position, category, probability and area size of the target in the RS image are utilized as input features $X^{(l)}$ of GCN after embedding. $\widetilde{A} = A + I$, $A$ is the adjacency matrix constructed from the distances among targets in the RS image, and $I$ is the identity matrix. $\widetilde{D}$ is the degree matrix [46] of $A$, which is used to normalize $A$. The distance $d_A$ between each two targets is defined to obtain an effective adjacency matrix

$$d_A(x, y) = e^{-||x-y||_2^2}\left(1 - ||x-y||_2^2\right) \quad (12)$$

where $x$ and $y$ are the coordinates of the two targets in the RS image. $||x||_2^2$ represents to calculate the length of the vector $x$. $d_A$ aims to weaken the relationship between distant targets as much as possible, while strengthening the relationship between close targets. By modeling local objects, GCN obtains scene semantics according to the object relationship.

For several instances in the natural scene, the location relationship between these instances can be automatically obtained using GCN. However, due to the redundant characteristic of the RS targets, directly using GCN to characterize each instance in the RS image will inevitably lead to the network being unable to focus on a useful feature representation. And the network has to deal with a large number of targets, resulting in excessive time consumption during training and inference [47]. Moreover, large targets in RS images are more likely to be described than small target. Inspired by this, the representation matrix and adjacency matrix are optimized to allow the network to focus more on salient instance features.

Firstly, if there are a large number of repeated targets in an RS image, such as a parking lot with hundreds of cars, we only need to let the model know that there are many cars instead of more specific details. Two objects that are close in distance and have the same type, need to be filtered and integrated into a representation that can reflect the information of the area. Define the critical similarity threshold $\breve{s}$ and filtration factor $\breve{f}(x, y)$ between the objects $x$ and $y$, that is

$$\breve{f}(x, y) = \Upsilon_{d_A > \breve{s}}\Upsilon_{\text{cate}_x == \text{cate}_y} \quad (13)$$

where $\Upsilon_{\text{status}}$ is an indicator function assigned to 1 if status is established, and $\text{cate}_x$ represents the category of the object $x$. When $\breve{f}(x, y)$ is 1, the target $x$ and $y$ need to be integrated to reduce the pressure of the data processing and increase the model's attention on valuable features. The following strategies are utilized when integrating the objects:

$$px_z = \text{Ave}(px_x, px_y), \quad py_z = \text{Ave}(py_x, py_y) \quad (14)$$

$$\text{area}_z = \text{area}_x \oplus \text{area}_y \quad (15)$$

$$\text{prob}_z = \text{Max}(\text{prob}_x, \text{prob}_y) \quad (16)$$

where $z$ is the target after integration. $(px_x, py_x)$ is the center coordinate of the target $x$. $\text{Ave}(x, y)$ and $\text{Max}(x, y)$ respectively represents the average and the maximum of $x$ and $y$. $\text{area}_x$ indicates the proportion of the total pixels occupied by the target $x$. $\text{prob}_x$ represents the probability that the target $x$ is the ground truth. $\oplus$ denotes vector sum. This strategy is used to filter every pair of targets in the RS image, which relieves the redundancy of local targets and greatly reducing the computational complexity.

Furthermore, we optimize the adjacency matrix to enhance the model's attention to salient targets. Objects with large areas tend to have a higher probability of being described, while small objects have a lower probability to be described. To enable the model to have a higher focus on the former, we boost the adjacency relationship of objects with large areas. Since the adjacency matrix $A$ contains the relation between the object pairs, we start directly with the adjacency matrix to optimize the representation. Specifically, the area that reflects the attention degree is leveraged to rank targets

$$\{\text{obj}_1^{\text{area}}, \ldots, \text{obj}_m^{\text{area}}, \ldots, \text{obj}_j^{\text{area}}\} = R_{\text{area}}\left(\{\text{obj}\}_1^j\right) \quad (17)$$

among them, $R_{\text{area}}$ indicates ranking objects based on area size. $\text{obj}_m^{\text{area}}$ represents the $m$th largest target in terms of area size. The area ranking is used to generate the relationship gain between the $m$th and $n$th target, we define the improved adjacency distance $\widehat{d_A}$

$$\widehat{d_A}\left(\text{obj}_m^{\text{area}}, \text{obj}_n^{\text{area}}\right) = \eth\,e^{-\sqrt{mn}}d_A\left(\text{obj}_m^{\text{area}}, \text{obj}_n^{\text{area}}\right) \quad (18)$$

where $\eth$ represents the relationship boost factor. Formula (18) aims to enhance the relationship of targets with a larger area and weaken the relationship of targets with a smaller area. The improved adjacency relationship further modifies the focus position of the model, and enables to obtain more effective local representations for visual encoding.

### C. Multi-Level Information Dynamic Fusion (MIDF)

In this subsection, we propose a multi-level information fusion method based on the global and local characteristics of the RS field.

Although global and local features can be acquired, generating the effective representation of RS images through these two features is still an urgent problem. On the one hand, the information at two levels needs to complement each other. Local features containing target adjacency information are often needed to be used to correct global information that contain redundant representations. At the same time, as local features contain less information, they need to be supplemented by global features with a wealth of information. On the other hand, appropriate weights are required to realize the summation of these two features. Although manually adjusting the weights of the two features can achieve well-retrieval accuracy, the parameter selection makes this method time-consuming and labor-intensive. If the model can dynamically generate corresponding weights according to the global and local features, it will significantly improve the visual representation of RS images. To solve aforementioned two problems, we propose the MIDF module and manage to fuse the multi-level features. MIDF performs a secondary representation of multi-level information to optimize features, and generates dynamic variable weights to achieve better visual representation.

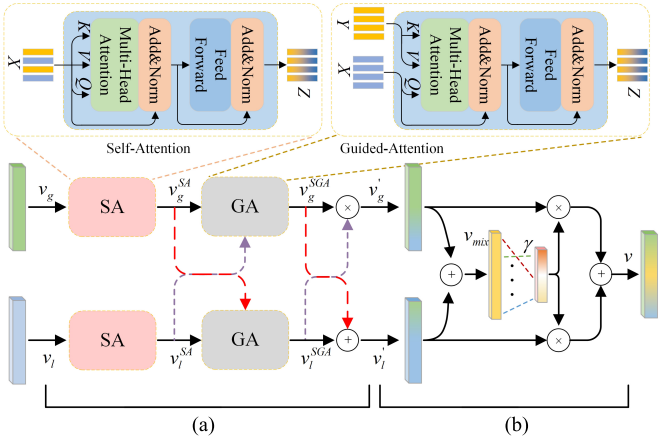

Fig. 3. Proposed multi-level information dynamic fusion module. The method falls into two stages of feature retransformation and dynamic fusion. MIDF first uses SA and GA modules to retransform features, then uses global information to supplement local information and leverages the latter to correct the former. Further dynamic fusion of multi-level features is accomplished through the devised dynamic fusion module. (a) Feature retransformation. (b) Dynamic fusion.

Specifically, we first re-characterize the two features to enable better interaction between global and local features. Inspired by [48], [49], we use self-attention (SA) and guided-attention (GA) to optimize the generated multi-level information. The SA module discovers the correlation between the information by performing an internal similarity calculation on the input features $X$. While the GA module outputs the attended features $Z$ for $X$ guided by features $Y$ by calculating the similarity of features $X$ and $Y$. As shown in Fig. 3, after passing the two features through SA module, a GA module is utilized to enable the two to interact

$$v_g^{\text{SA}} = \text{SA}(v_g), \quad v_l^{\text{SA}} = \text{SA}(v_l) \tag{19}$$

$$v_g^{\text{SGA}} = \text{GA}(v_g^{\text{SA}}, v_l^{\text{SA}}), \quad v_l^{\text{SGA}} = \text{GA}(v_l^{\text{SA}}, v_g^{\text{SA}}) \tag{20}$$

where $\text{SA}(x)$ indicates processing feature $x$ with the self-attention module, and $\text{GA}(x, y)$ represents using feature $y$ to guide the representation of feature $x$. $v_g^{\text{SA}}$ and $v_l^{\text{SA}}$ indicate global and local obtained after processing by the SA module, $v_g^{\text{SGA}}$ and $v_l^{\text{SGA}}$ indicate global and local features after feature interaction. Further, to use local features to correct global features, while making global features supplement local features, we define the following operations:

$$v_g' = v_g^{\text{SGA}} \otimes \sigma(v_l^{\text{SGA}}), \quad v_l' = v_l^{\text{SGA}} \oplus v_g^{\text{SGA}} \tag{21}$$

among them, $\otimes$ represent dot product. $v_g'$ and $v_l'$ are the global and local features after information interaction. Formula (21) enables the local features generate a mask to filter the global features, and makes the global features directly supplement the local features at the same time.

Although the multi-level features perform better after interaction, there is still a need to obtain a hybrid representation of visual features. To generate dynamic weights based on information at different levels, we first superimpose these two features to obtain the mixed visual information $v_{\text{mix}}$. Then the learnable dynamic weights $\gamma$ are obtained by linear transformation on $v_{\text{mix}}$, which further is used to generate the fused features $v$

$$v_{\text{mix}} = v_g' \oplus v_l' \tag{22}$$

$$\gamma_1, \gamma_2 = \text{Softmax}\big(\sigma(v_{\text{mix}} W_\alpha) W_\beta\big) \tag{23}$$

$$v = \big(\gamma_1 \otimes v_g'\big) \oplus \big(\gamma_2 \otimes v_l'\big) \tag{24}$$

where $W_\alpha$ and $W_\beta$ are the weight matrices and $\text{Softmax}(x)$ denotes using softmax function to activate feature $x$. The proposed MIDF module complements multi-level information through feature re-transformation and dynamic fusion steps to obtain a more competitive feature representation.

### D. Multivariate Rerank (MR)

In this subsection, a post-processing method for retrieval tasks has been proposed, which optimizes retrieval results by combining multiple variable information in secondary sorting.

In the traditional retrieval task, the similarity calculation between $M$ query texts and $N$ queried images will produce a similarity matrix $S_{\text{src}}$ with size $M \times N$. The top $k$ results of $S_{\text{src}}[t]$ after sorting are the top $k$ candidate images retrieved by text $t$ in the dataset. The retrieval principle is still followed when using RS image for text retrieval. Although the correctness of this retrieval method is undoubted, it ignores the internal relationship between bidirectional retrieval, which is crucial to further improve retrieval accuracy [50]. More precisely, once the text and image match, they must be mutually retrievable. Based on this, Wang *et al.* [50] proposed a cross-modal rerank algorithm, which uses $k$ candidates to perform a reverse search, and obtains the final retrieval result according to the ranking position in the reverse retrieval result. Even if the cross-modal rerank algorithm can improve the retrieval performance to a certain extent, the method does not further fully consider the information of the similarity matrix.

To fully integrate a variety of useful information in bidirectional retrieval, as shown in Fig. 4, a MR algorithm is proposed. To make full use of the similarity matrix $S_{\text{src}}$, the basic idea is to use candidates for reverse search and to optimize the retrieval results by considering multiple ranking factors. The devised algorithm integrates the ranking information of $i2t$ and $t2i$ in the original similarity matrix, while adding a significant component to further enhance the reranking accuracy. Next, we will take $i2t$ as an example to illustrate the algorithm of MR in detail.

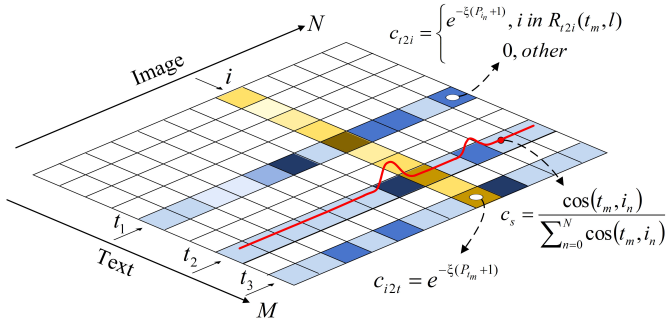

Fig. 4. Proposed multivariate rerank algorithm. To make full use of the similarity matrix, we use $k$ candidates for reverse search and to optimize the similarity results by considering multiple ranking factors. The figure shows an illustration of multivariate rerank when $k = 3$ and using image $i$ for retrieval.

First, $R_{i2t}(i, k)$ is defined as a query in the similarity matrix $S_{src}$ using $i$, and $k$ denoting the top $k$ nearest neighbors

$$t_1, \ldots, t_m, \ldots, t_k = R_{i2t}(i, k) \tag{25}$$

where $t_m$ denotes the $m$th text which is most similar to the query image. At this point, we have initially obtained the ranking position $P_{t_m} \in (0, 1, \ldots, M)$ of each candidate text. To fully utilize this information, we define the $i2t$ component $c_{i2t}$

$$c_{i2t} = e^{-\xi(P_{t_m} + 1)} \tag{26}$$

where $\xi$ is the ranking coefficient. The purpose of this operation is to normalize the $i2t$ ranking information, and the top-ranked text has a higher $i2t$ component.

Next, we use the candidates obtained in the previous step to perform reverse search, respectively. When using the text $t_m$ to query, $R_{t2i}(t_m, l)$ is defined as a query in the similarity matrix $S_{src}$ using $t_m$

$$i_1, \ldots, i_n, \ldots, i_l = R_{t2i}(t_m, l) \tag{27}$$

where $l$ represents the top $l$ nearest neighbors and $i_n$ denotes the $n$th nearest neighbor image. For the $k$ nearest neighbor texts corresponding to each retrieved image, $l$ nearest neighbor images for reverse retrieval can be obtained. At the same time, we can also get the position $P_{I_n}$ if the query image $I$ is in the $L$ nearest neighbor images. Automatically, we define the $t2i$ component as

$$c_{t2i} = \begin{cases} e^{-\xi(P_{i_n} + 1)}, & i \text{ in } R_{t2i}(t_m, l) \\ 0, & \text{other} \end{cases} \tag{28}$$

$c_{t2i}$ represents the secondary similarity confirmation when reverse retrieval is performed, which can be used to correct $c_{i2t}$.

Further, we define significance components $c_s$ to quantify the degree of confirmation on similarity predicted by the model. For candidate text $t_m$ in reverse retrieval, the higher the proportion of similarity with $i_n$ to the similarity with all images, the higher the degree of certainty. For this purpose, we calculate the ratio and regard it as a confidence component. In particular

$$c_s = \frac{\cos(t_m, i_n)}{\sum_{n=0}^{N} \cos(t_m, i_n)} \tag{29}$$

$c_s$ calculates the confidence of the model on the source similarity, which is utilized as a weighting term for the final similarity.

Finally, we weight the three reranking components to obtain the similarity after MR $S_{mr}$

$$S_{mr} = c_{i2t} + w_{c_1} c_{t2i} + w_{c_2} c_s \tag{30}$$

$S_{mr}$ comprehensively considers the results of forward and reverse search, and combines the model's confirmation of the initial similarity to do the secondary ranking. $S_{mr}$ takes more factors into account than the original similarity, which definitely leads to a more fine-grained ranking result. The experiment section shows the powerful performance of MR algorithm.

### E. Objective Function

With the development of multimodal feature alignment, triplet loss has been called one of the mainstream loss functions in the field of multimodal feature matching. For the samples in one batch during training, triplet loss makes the distance between paired samples smaller than the distance between unpaired samples by a fixed value

$$L(I, T) = \sum_{\widehat{T}} \left[ \varepsilon - \cos(I, T) + \cos\left(I, \widehat{T}\right) \right]_+ \\ + \sum_{\widehat{I}} \left[ \varepsilon - \cos(I, T) + \cos\left(\widehat{I}, T\right) \right]_+ \tag{31}$$

where $\varepsilon$ represents the minimum margin, $[x]_+ \equiv \max(x, 0)$. $(I, T)$ is a paired sample pair. $\widehat{T}$ is the text that not paired with the image $I$, and $\widehat{I}$ is the RS image not paired with the text $T$. The triplet loss aims to make the distance between the anchor and the negative sample as far as possible, and make it as close as possible to the positive sample. Follow [14], [51], we use the triplet loss as our objective function in our work.

### F. Training and Inference Procedure

The training and infernece procedures of the proposed GaLR are shown in Algorithm 1 and Algorithm 2, respectively.

During the training stage, RS image-text $D_{tr}$ is needed. We use ppyolo [52] with a backbone of ResNet-50 as the object detector, and utilize it for object extraction of RS images after pre-training the model on the DOTA dataset. Simultaneously, the multiple sub-modules of GaLR are initialized. For each batch $\mathbf{I}, \mathbf{T} \in \mathbf{D}_{tr}$ of the training set, we first use CNN with MVSA to extract global features $v_g$. Then the detector is used to obtain the targets in the RS image, which are processed into adjacency matrix $M_A$ and representation matrix $M_R$ through the relation construction module with DREA mechanism. GCN extracts the features of the $M_A$ and $M_R$, and we denote it as the local feature $v_l$. Further, MIDF module

**Algorithm 1** Training Procedure of the Proposed GaLR

**Require:**

RS image-text train dataset $\mathbf{D_{tr}} = \{\{I_1, T_1\}, \{I_2, T_2\}, \ldots\}$ ($I$ is labeled image, $T$ is corresponding sentence),
ppyolo model $\{obj\}_1^j = Detect(I; \vartheta)$ pre-trained on the DOTA dataset,
CNN with MVSA module $v_g = \mathcal{C}_{mask}(I; \theta_1)$,
Relation construction $M_A, M_R = RC_{\text{DREA}}(\{obj\}_1^j)$,
GCN module $v_l = \mathcal{G}(M_A, M_R; \theta_2)$,
MIDF module $v = \text{MIDF}(v_g, v_l; \theta_3)$,
GRU module $t = \text{GRU}(T; \theta_4)$

**Repeat until convergence:**

1: **for** each batch $\mathbf{I}, \mathbf{T} \in \mathbf{D_{tr}}$ **do**
   **Visual feature extraction**
2:   $v_g = \mathcal{C}_{mask}(I; \theta_1)$
3:   $\{obj\}_1^j = Detect(I; \vartheta)$
4:   $M_A, M_R = RC_{\text{DREA}}(\{obj\}_1^j)$
5:   $v_l = \mathcal{G}(M_A, M_R; \theta_2)$
6:   $v = \text{MIDF}(v_g, v_l; \theta_3)$
   **Text feature extraction**
7:   $t = \text{GRU}(T; \theta_4)$
   **Calculate the triplet loss**
8:   $l_{tpt} = L(v, t)$
9:   **Update** $\theta_1, \theta_2, \theta_3, \theta_4$ **by** $l_{tpt}$
10: **end for**
11: **return** $\mathcal{C}_{mask}$, $\mathcal{G}$, **MIDF, GRU**

**Algorithm 2** Inference Procedure of the Proposed GaLR

**Require:**

RS image-text test dataset $\mathbf{D_{ir}} = \{\{I_1, T_1\}, \{I_2, T_2\}, \ldots\}$ ($I$ is labeled image, $T$ is corresponding sentence),
GaLR model $v, t = \text{GaLR}(I, T; \theta)$,
Initialize the similarity matrix $S$,
Multivariate rerank $S_{mr} = \text{MR}(S)$

**Calculate image-text similarity matrix:**

1: **for** each batch $\mathbf{I}, \mathbf{T} \in \mathbf{D_{ir}}$ **do**
   **Visual feature extraction**
2:   $v, t = \text{GaLR}(I, T; \theta)$
3:   $S_b = cos(v, t)$
4:   Append $S_b$ to $S$
5: **end for**
6: **return** $S$

**Post-processing using multivariate rerank:**

7:   $S_{mr} = \text{MR}(S)$
8: **return** $S_{mr}$

fuses global features $v_g$ and local features $v_l$ to obtain visual representation $v$. For text representation, the bidirectional GRU directly extracts the features $t$. The triplet loss function calculates the loss produced by the distances of single-modal features $v$ and $t$, and utilizes it to optimize the submodules of GaLR.

In the inference stage, for each batch $\mathbf{I}, \mathbf{T} \in \mathbf{D_{ir}}$ of the test set, we first use the GaLR model to obtain the single-modal representation $v$ and $t$. The cosine similarity is leveraged to

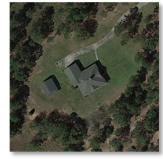

Cap1: two roads pass through the green, trees are on both sides of them.

Cap2: a house are on the light green ground and two cars are parked at the gate.

Cap3: many sparsely distributed green trees are around two gray buildings.

Cap4: many sparsely distributed green trees are around two gray buildings.

Cap5: two roads pass through the green, trees are on both sides of them.

(a)

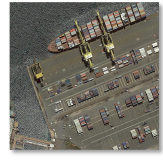

Cap1: a red dry freighter full of cargo is on the shore unloading.

Cap2: a ship full of cargo docked by the deep blue sea.

Cap3: the yellow unloader at the port is unloading cargo from the ship.

Cap4: two yellow unloaders on the gray port are unloading.

Cap5: the grey port was neatly packed with goods.

(b)

Fig. 5. For each sample in remote sensing text-image datasets, an RS image and five corresponding sentences are included. Samples from RSICD and RSITMD are shown in this figure, respectively. (a) RSICD sample. (b) RSITMD sample.

calculate the similarity $S_b$ of the single-modal representation in one batch, and then $S_b$ is appended to the similarity matrix $S$ of the entire image-text test set. The proposed post-processing of MR is utilized to make $S$ have a secondary optimization, and then the final similarity matrix $S_{mr}$ is obtained.

## IV. EXPERIMENTS RESULTS AND ANALYSIS

In this section, we first introduce the dataset, evaluation indicators, and experiment details. Hereafter, we conduct comparisons and analyses on several RS text-image datasets to verify the effectiveness of the devised GaLR model. Further, plenty of ablation experiments have been carried out to systematically explore the reasons for the excellent performance of the GaLR model and MR algorithm. Finally, the visualization experiment and the comparison of retrieval time are presented.

### A. Dataset and Evaluation Metrics

Two commonly used RS text-image datasets are used in this experiment: RSICD and RSITMD. As shown in Fig. 5, each sample in the above dataset contains an RS image and five corresponding sentences. RSICD [23] has 10921 samples with the image size of $224 \times 224$, which is the largest RS text-image dataset in sample numbers. RSITMD [14] has 4743 samples and the image size is $256 \times 256$, which has more fine-grained representation in text than RSICD. In experiments, we follow the data partitioning approach of Yuan *et al.* [14] and use 80%, 10%, and 10% of the dataset as the training set, validation set and test set, respectively.

We use $R@k$ and $mR$ [53] as evaluation criteria to assess the model performance. $R@k$ aims to calculate the proportion of ground truth in the top $k$ samples recalled. In the experiment, $k$ is taken as 1, 5, and 10 to more reasonably evaluate the results. $mR$ is then leveraged to represent the average of multiple $R@k$, which reflects the overall performance of the model more intuitively.

### B. Implementation Details

All experiments in this article are conducted on a single NVIDIA RTX 3090 GPU. For different image sizes in different datasets, we scale them uniformly to $256 \times 256$ pixels and feed them into the network. We have performed a series of data enhancements such as rotation and flip to improve the robustness of the model. The representation dimension of the word vector is set to 300. The embedding space for both images and text is 512. The margin is restricted to 0.2 for the triplet loss calculation. When optimizing the adjacency matrix, the critical similarity threshold and the relationship boost factor are set to 0.8 and 1.15, respectively. We use the Adam optimizer with the triplet loss to train the network 70 epochs, and the batchsize is 100. The initial learning rate is 1e-4 and decays by 0.7 every 20 epochs. To make the experiment more convincing, for each set of experiments, we conduct the same training and verification five times to obtain an average result.

TABLE I
COMPARISONS OF RETRIEVAL PERFORMANCE ON RSICD AND RSITMD TESTSET

| Approach | RSICD dataset | | | | | | | RSITMD dataset | | | | | | |
|---|---|---|---|---|---|---|---|---|---|---|---|---|---|---|
| | Sentence Retrieval | | | Image Retrieval | | | mR | Sentence Retrieval | | | Image Retrieval | | | mR |
| | R@1 | R@5 | R@10 | R@1 | R@5 | R@10 | | R@1 | R@5 | R@10 | R@1 | R@5 | R@10 | |
| VSE++ | 3.38 | 9.51 | 17.46 | 2.82 | 11.32 | 18.10 | 10.43 | 10.38 | 27.65 | 39.60 | 7.79 | 24.87 | 38.67 | 24.83 |
| SCAN t2i | 4.39 | 10.90 | 17.64 | 3.91 | 16.20 | 26.49 | 13.25 | 10.18 | 28.53 | 38.49 | 10.10 | 28.98 | 43.53 | 26.64 |
| SCAN i2t | 5.85 | 12.89 | 19.84 | 3.71 | 16.40 | 26.73 | 14.23 | 11.06 | 25.88 | 39.38 | 9.82 | 29.38 | 42.12 | 26.28 |
| CAMP-triplet | 5.12 | 12.89 | 21.12 | 4.15 | 15.23 | 27.81 | 14.39 | 11.73 | 26.99 | 38.05 | 8.27 | 27.79 | 44.34 | 26.20 |
| CAMP-bce | 4.20 | 10.24 | 15.45 | 2.72 | 12.76 | 22.89 | 11.38 | 9.07 | 23.01 | 33.19 | 5.22 | 23.32 | 38.36 | 22.03 |
| MTFN | 5.02 | 12.52 | 19.74 | 4.90 | 17.17 | 29.49 | 14.81 | 10.40 | 27.65 | 36.28 | 9.96 | 31.37 | 45.84 | 26.92 |
| LW-MCR(b) | 4.57 | 13.71 | 20.11 | 4.02 | 16.47 | 28.23 | 14.52 | 9.07 | 22.79 | 38.05 | 6.11 | 27.74 | 49.56 | 25.55 |
| LW-MCR(d) | 3.29 | 12.52 | 19.93 | 4.66 | 17.51 | 30.02 | 14.66 | 10.18 | 28.98 | 39.82 | 7.79 | 30.18 | 49.78 | 27.79 |
| AMFMN-soft | 5.05 | 14.53 | 21.57 | **5.05** | 19.74 | 31.04 | 16.02 | 11.06 | 25.88 | 39.82 | 9.82 | 33.94 | 51.90 | 28.74 |
| AMFMN-fusion | 5.39 | 15.08 | 23.40 | 4.90 | 18.28 | 31.44 | 16.42 | 11.06 | 29.20 | 38.72 | 9.96 | 34.03 | 52.96 | 29.32 |
| AMFMN-sim | 5.21 | 14.72 | 21.57 | 4.08 | 17.00 | 30.60 | 15.53 | 10.63 | 24.78 | 41.81 | **11.51** | 34.69 | **54.87** | 29.72 |
| GaLR w/o MR | 6.50 | 18.91 | 29.70 | **5.11** | **19.57** | 31.92 | 18.62 | 13.05 | 30.09 | **42.70** | 10.47 | 36.34 | 53.35 | 31.00 |
| GaLR with MR | **6.59** | **19.85** | **31.04** | 4.69 | 19.48 | **32.13** | **18.96** | **14.82** | **31.64** | 42.48 | 11.15 | **36.68** | 51.68 | **31.41** |

### C. Comparisons With the SoTA Methods

We compare the proposed model with the following multiple progressive cross-modal retrieval models on two RS text-image datasets.

- *VSE++ [41]:* VSE++ extracts image and text features by CNN and GRU, respectively, and directly uses triplet loss to optimize them.
- *SCAN [51]:* SCAN enhances the VSE++ method by aligning local features, which are extracted by object detector.
- *CAMP [54]:* CAMP uses a message passing mechanism to dynamically control the flow of cross-modal information, and then obtain the final results by cosine similarity.
- *MTFN [50]:* MTFN leverages a fusion of multiple features to calculate cross-modal similarity end-to-end.
- *AMFMN [14]:* AMFMN utilizes filtered visual features to guide the representation of text features through an asymmetrical architecture.
- *LW-MCR [15]:* Lightweight multi-scale crossmodal retrieval method (LW-MCR) takes advantage of methods such as knowledge distillation and contrast learning for lightweight retrieval models.

In the above experiments, ResNet-18 [55] is used as the image feature extractor for all models except LW-MCR to compare the experiment results fairly. For the first four methods, we migrated the author's published code to the RS image-text dataset, fine-tuned the specific parameters, and performed cross-validation to ensure model performance with a consistent model backbone. For the latter two methods, we use the results in literature [15].

We list two GaLR methods to compare with the above methods.

- *GaLR w/o MR:* GaLR model without MR.
- *GaLR with MR:* GaLR model with MR.

Table I shows the experiment results on different datasets, we can get the following conclusions.

- On the RSICD, the performance of the GaLR is particularly impressive. For the GaLR model without the MR algorithm, the *mR* score reaches 18.62, which surpasses the most outstanding models by 2.2 points. After adding the MR mechanism, the score of the optimized GaLR has been improved to 18.96. The MR mechanism further increases the retrieval results through secondary sorting. RSICD is a large RS image-text dataset, and the performance of GaLR on this dataset strongly illustrates the effectiveness of the proposed method.
- The GaLR also still performs well on the RSITMD dataset. The accuracy of GaLR without MR mechanism has reached 31.00, which is 1.28 points ahead of AMFMN. After MR, GaLR achieves the first position among multiple indicators and wins SoTA performance with the *mR* metric of 31.41. The excellent performance on the RSITMD once again validates the superiority of the proposed method in the RSCTIR task.

### D. Performance Analysis of DREA Mechanism

A large number of ablation experiments have been done to study the effect of DREA mechanism on retrieval performance. We first obtain local features through the unimproved representation (LF w/o DREA), and then add the DREA mechanism (LF with DREA) to compare the effect with the former. To fully illustrate the problem, we remove the proposed MIDF module and observe the change of retrieval performance by adjusting the weights of global and local features. The final experiment results are shown in Table II.

TABLE II
PERFORMANCE COMPARISON WHEN USING DREA MECHANISM ON RSITMD TESTSET

| Ablation Model | | $w_g$ | $w_l$ | Sentence Retrieval | | | Image Retrieval | | | mR | Growth |
|---|---|---|---|---|---|---|---|---|---|---|---|
| | | | | R@1 | R@5 | R@10 | R@1 | R@5 | R@10 | | |
| LF with DREA | m1 | 1.0 | 0.0 | 12.17 | 27.43 | 38.35 | 9.10 | 31.11 | 46.61 | 27.46 | - |
| | m2 | 0.8 | 0.2 | 11.06 | 28.83 | 40.71 | 10.72 | 34.99 | 52.05 | 29.73 | - |
| | m3 | 0.6 | 0.4 | 13.27 | 28.76 | 40.04 | 10.86 | 35.10 | 49.99 | 29.67 | - |
| | m4 | 0.5 | 0.5 | 11.06 | 28.54 | 42.48 | 9.82 | 34.60 | 52.92 | 29.90 | - |
| | m5 | 0.4 | 0.6 | 10.03 | 27.29 | 40.27 | 10.12 | 36.33 | 55.18 | 29.87 | - |
| | m6 | 0.2 | 0.8 | 6.64 | 20.87 | 32.01 | 6.00 | 25.75 | 45.40 | 22.78 | - |
| | m7 | 0.0 | 1.0 | 2.36 | 6.56 | 11.06 | 1.92 | 8.72 | 15.01 | 7.61 | - |
| LF w/o DREA | m1 | 1.0 | 0.0 | 12.17 | 27.43 | 38.35 | 9.10 | 31.11 | 46.61 | 27.46 | - 0.00 |
| | m2 | 0.8 | 0.2 | 11.87 | 31.05 | 41.88 | 10.82 | 35.83 | 52.80 | 30.71 | ↑ 0.98 |
| | m3 | 0.6 | 0.4 | 12.32 | 29.57 | 42.77 | 10.97 | 35.93 | 53.54 | 30.85 | ↑ 1.18 |
| | m4 | 0.5 | 0.5 | 11.96 | 28.54 | 42.48 | 8.94 | 35.66 | 55.76 | 30.52 | ↑ 0.62 |
| | m5 | 0.4 | 0.6 | 11.72 | 28.76 | 40.27 | 9.90 | 36.61 | 56.14 | 30.57 | ↑ 0.70 |
| | m6 | 0.2 | 0.8 | 5.24 | 18.36 | 29.06 | 5.07 | 24.29 | 43.36 | 20.90 | ↓ 1.88 |
| | m7 | 0.0 | 1.0 | 1.17 | 3.39 | 5.68 | 1.15 | 4.85 | 8.89 | 4.19 | ↓ 3.42 |

TABLE III
CONTRAST EXPERIMENT OF RERANK

| Optim Method | Sentence Retrieval | | | Image Retrieval | | | mR |
|---|---|---|---|---|---|---|---|
| | R@1 | R@5 | R@10 | R@1 | R@5 | R@10 | |
| Source | 13.05 | 30.09 | **42.70** | 10.47 | 36.34 | **53.35** | 31.00 |
| Rerank | **14.82** | 30.75 | 42.26 | **11.19** | **36.77** | 51.46 | 31.21 |
| Multivariate Rerank | **14.82** | **31.64** | 42.48 | 11.15 | 36.68 | 51.68 | **31.41** |

When the global weight $w_g$ proportion is 1.0, the network degenerates into the visual coding branch of AMFMN and we regard it as the baseline model. For the LF w/o DREA, when the local feature proportion is gradually increased to 0.6, the retrieval accuracy has also improved from 27.46 at baseline to 29.87. The change in accuracy before and after the addition of local features verifies the importance of local features in optimizing retrieval performance. However, when the proportion of local features continues to be increased, the retrieval accuracy shows a gradual decreasing trend. The augmentation of local features causes the model to rely excessively on local information and gradually ignore global features, which enables the model to have a better understanding of the overall RS image.

When the DREA is added for local feature representation, the experiment results again validate the above analysis. Further, we analyze the impact of DREA on retrieval performance under specific weights. When the local feature is between 0.2 and 0.6, the improvement in retrieval performance of LF with DREA is significant. In this case, the model can perceive more salient targets in the RS image, and combine with global features to obtain a more accurate retrieval accuracy. While when the local feature component continues to rise, the performance of retrieval using LF with DREA decreases compared to LF w/o DREA. The reason is that DREA filters out a large number of targets, which leads to a lack of information which cannot be compensated by small-component global features. Despite this, LF with DREA still achieves the highest retrieval accuracy at $w_l$ of 0.4, which means that the lack of performance is acceptable in the case of a high degree of local features.

### E. Rerank Versus MR

In this subsection, we conduct ablation experiments of MR to verify the effectiveness of the proposed secondary optimization algorithm. We optimize the different similarity matrices on the RSITMD dataset, which are generated by GaLR with different parameters, to demonstrate the generality and stability of the method. We also regard the rerank method mentioned in [50] as a control group to explore the advantages of the proposed MR.

The source retrieval results, rerank results, and MR results averaged by multiple similarity matrices are shown in Table III. In this experiment, $k$ is 25, and the values of $w_{c_1}$ and $w_{c_2}$ in the MR algorithm are set to 0.5 and 1.25. Before optimization, the metric $mR$ when using the source similarity matrix for testing is 31.00, which is obtained by averaging the results of multiple similarity matrices. Then, we utilize the rerank method to reorder the similarity, and the retrieval accuracy has been improved by 0.21 points. Although the $mR$ after rerank raise a little, because the rerank mechanism only uses the reverse ranking as a candidate, a large part information of the similarity matrix is ignored. Further, we use MR to optimize the original similarity matrix. Compared with the optimization algorithm using the rerank mechanism, the proposed MR mechanism mines the information in the similarity matrix more diversely. MR method improves the retrieval accuracy to a greater extent and raises the retrieval accuracy on RSITMD to 31.41.

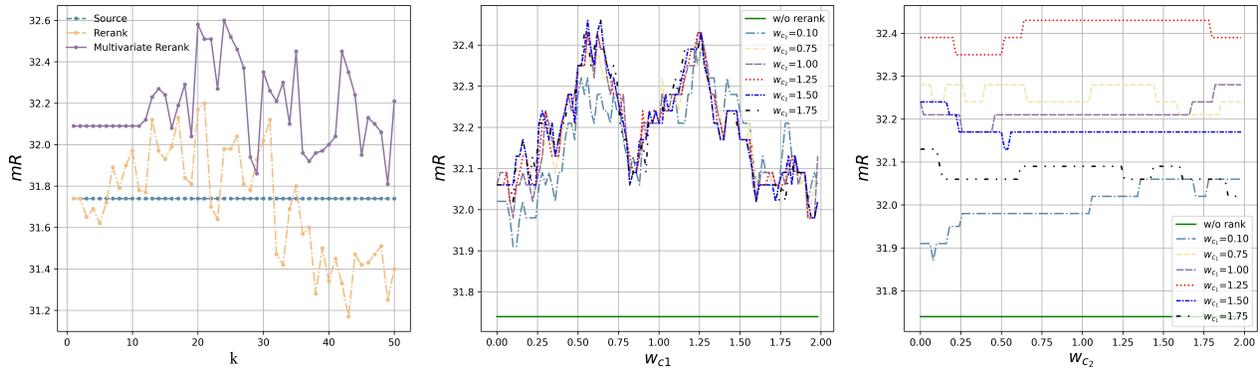

Fig. 6. Variable analysis experiments of multivariate ranking algorithms for $k$, $W_{c_1}$, and $W_{c_2}$ on RSITMD testset. From left to right are: (a) Comparison of the original results, the results of rerank, and the results after MR algorithm when $k$ varies. (b) Effect of $W_{c_1}$ on retrieval results when fixing $W_{c_2}$. (c) Effect of $W_{c_2}$ on retrieval results when fixing $W_{c_1}$.

TABLE IV
Ablation Experiment for Different Structures of GaLR on RSITMD Testset

| Ablation Model | Global Rep | | Local Rep | | Fusion Rep | | Multivariate Rerank | Sentence Retrieval | | | Image Retrieval | | | mR |
|---|---|---|---|---|---|---|---|---|---|---|---|---|---|---|
| | CNN | MVSA | GCN | DREA | ADD | MIDF | | R@1 | R@5 | R@10 | R@1 | R@5 | R@10 | |
| g1 | ✓ | | | | | | | 10.38 | 27.65 | 39.6 | 7.79 | 24.87 | 38.67 | 24.83 |
| g2 | ✓ | ✓ | | | | | | 12.17 | 27.43 | 38.35 | 9.10 | 31.11 | 46.61 | 27.46 |
| l1 | | | ✓ | | | | | 2.36 | 6.56 | 11.06 | 1.92 | 8.72 | 15.01 | 7.61 |
| l2 | | | ✓ | ✓ | | | | 1.17 | 3.39 | 5.68 | 1.15 | 4.85 | 8.89 | 4.19 |
| f1 | ✓ | ✓ | ✓ | | ✓ | | | 11.06 | 28.54 | 42.48 | 9.82 | 34.60 | 52.92 | 29.90 |
| f2 | ✓ | ✓ | ✓ | ✓ | ✓ | | | 12.32 | 29.57 | **42.77** | 10.97 | 35.93 | **53.54** | 30.85 |
| f3 | ✓ | ✓ | ✓ | ✓ | | ✓ | | 13.05 | 30.09 | 42.70 | 10.47 | 36.34 | 53.35 | 31.00 |
| m1 | ✓ | ✓ | ✓ | ✓ | | ✓ | ✓ | **14.82** | **31.64** | 42.48 | **11.15** | **36.68** | 51.68 | **31.41** |

To analyze the influence of the parameters $k$, $w_{c_1}$, and $w_{c_2}$ on the experiment results, we conduct a variable analysis for the three parameters on RSITMD. To show the effectiveness of the MR algorithm more intuitively, we select a typical similarity matrix and visualize the change of $mR$ with parameters. Fig. 6(a) shows the variation of $mR$ indicator with the value of $k$. When $k$ varies, the rerank algorithm has better retrieval performance when $k$ takes values between 7 and 32. However, when $k$ continues to increase, the accuracy of the rerank algorithm drops sharply. For the MR algorithm, the $mR$ indicator also fluctuates considerably when $k$ changes, but is always above the result of the rerank algorithm. This experiment verifies the robustness of the proposed MR algorithm to the parameter $k$. Fig. 6(b) shows the effect of $w_{c_1}$ on the retrieval results when $k$ is 25 and fixing $w_{c_2}$. When $w_{c_1}$ is at 0.6 and 1.25, the MR algorithm achieves better retrieval accuracy. Similar to the variation of $k$, the retrieval accuracy of MR is always better than the original score regardless of the value of $w_{c_1}$ taken. Fig. 6(c) shows the effect of $w_{c_2}$ on $mR$ when $k$ is 25 and fixing $w_{c_1}$. Compared with $w_{c_1}$, the value of $w_{c_2}$ is less sensitive to the effect of retrieval accuracy, and the fluctuation of $w_{c_2}$ to $mR$ is within 0.2. MR mechanism comprehensively considers the bidirectional results in the similarity matrix and therefore obtains a more prominent retrieval result. The above experiments strongly validate the effectiveness of the proposed algorithm.

### F. Ablation Studies of Structures

In this subsection, we conduct detailed experiments to systematically analyze the proposed GaLR method. Eight control experiments are designed to explore the influence of each module on retrieval effectiveness.

- *g1:* Use only global visual features extracted by CNN.
- *g2:* Use only global visual features extracted by CNN with MVSA.
- *l1:* Use only local visual features extracted by GCN.
- *l2:* Use only local visual features extracted by GCN with DREA.
- *f1:* Use the sum of global features and local features extracted only using GCN.
- *f2:* Use the sum of global features and local features extracted using GCN with DREA.
- *f3:* Use MIDF module to fuse global and local features.
- *m1:* Use the MR algorithm to optimize the results obtained by the f3 model.

Among these models, the g1 and g2 models show the influence of the MVSA module on the experimental results. Control group of (l1, l2) and (f1, f2) demonstrates the improvement of the DREA mechanism on the retrieval results. Control group of (f2, f3) verifies the superiority of MIDF module. m1 shows the impact of the proposed MR algorithm in the final retrieval performance.

Table IV shows the results of five groups of experiments.

- Compared with g1, the mR indicator of the model g2 has increased by 2.63 points after adding the MVSA module.
- Compared with l1 and l2, when only local feature is utilized, DREA has a negative impact on performance. However, when the global feature is added, comparing l3 and l4, DREA improves the retrieval result by 0.95 points, which has been discussed in part D of the experiment.

| Task | Query | Method | Top 5 Results |
|------|-------|--------|---------------|
| Image → Text | 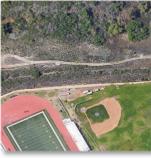 | **GaLR** | A green baseball field adjacent to the playground and Red Square. · The green baseball field is adjacent to the playground and the red building. · A playground is between a baseball field and a big building. · There is a long path in the field next to the red playground. · There are many trees on the playground. |
|  |  | AMFMN | The green playground around the red runway is a baseball field. · The football field is green and the grass around the trees is green. · A green baseball field adjacent to the playground and Red Square. · The green baseball field is adjacent to the playground and the red playground. · There is a large green space on the playground. |
|  |  | LW-MCR | There are a few trees around a plastic playground . · There is a long path in the field next to the red playground. · There is a baseball field beside the green amusement park around the red track. · There is a gray room next to a baseball field. · There are a few trees around a plastic playground . |
| Text → Image | There is a tennis court next to the football field and a blue building next to it. | **GaLR** | 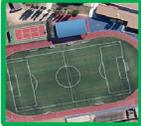 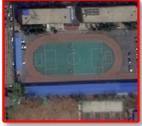 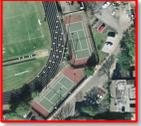 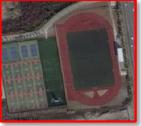 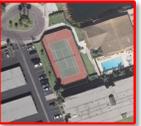 |
|  |  | AMFMN | 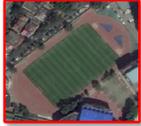 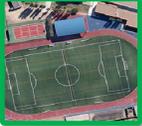 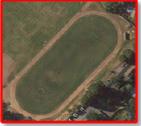 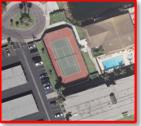 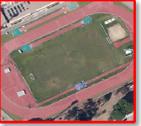 |
|  |  | LW-MCR | 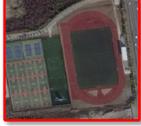 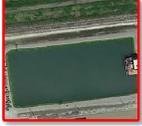 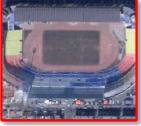 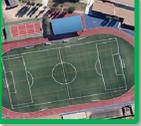 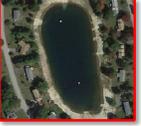 |

Fig. 7. Visualization of retrieval results. We respectively show the retrieval results of each model on the RSITMD testset when performing image-text and text-image retrieval tasks. In the top5 results, those surrounded by green boxes are ground truth, and those surrounded by red boxes are wrong results.

TABLE V
TIME-CONSUMING COMPARISON OF DIFFERENT METHODS

| Approach | VSE++ | SCAN | MTFN | CAMP | AMFMN | LW-MCR | GaLR |
|----------|-------|------|------|------|-------|--------|------|
| ET of RSICD (s) | 22.81 | 64.15 | 45.96 | 29.35 | 38.08 | **18.29** | 46.34 |
| ET of RSITMD (s) | 4.38 | 11.69 | 8.55 | 6.72 | 8.41 | **4.12** | 8.82 |
| IT (ms) | 3.25 | 4.85 | 4.60 | 4.06 | 3.51 | **2.83** | 4.20 |

- The comparison between g2 and l2 illustrates that the model using only global features is significantly better than the model using local features. The global features often contain all the information of RS image, while the local features filter out part of the information, which causes the poor retrieval performance. Although local features can reflect the relationships between salient objects, due to the high intra-class similarity of RS images, the model cannot perform a good analysis of the scene through only local features.

- The retrieval accuracy of f1 and f2 rises significantly when both types of information are leveraged simultaneously. This indicates that the two levels of features are indispensable when performing retrieval, and both are helpful to improve the retrieval accuracy.

- Although manually adjusting the weights of the two levels of information enables the model to achieve better retrieval accuracy, such an approach often requires considerable effort to adjust the trade-off parameters.

The f3 model performs a dynamic fusion of the different features. Even if the accuracy only rises by 0.15 points, there is no need to fine-tune the trade-off parameters during training process. On the one hand, MIDF makes it unnecessary to spend a lot of time on parameter adjustment during training, and on the other hand, it allows a small improvement in retrieval accuracy.

- The m1 model implements MR algorithm based on f3. In contrast to f3, the similarity matrix of m1 is considered from the perspective of multiple variables and thus the retrieval results are optimized.

### G. Visual Analysis of Recall Results

In this subsection, we visualize the typical experimental results to intuitively analyze the performance differences among various retrieval models. We have selected the AMFMN and LW-MCR models as the comparisons with our model. AMFMN uses a powerful feature extractor to extract the global features of RS image which can be directly used for retrieval. LW-MCR leverages a lightweight feature extractor and utilizes knowledge distillation to improve retrieval performance. In this experiment, to give a fair comparison, the multivariate rank algorithm is not used for secondary optimization of the similarity matrix. We conduct this experiment on the RSITMD dataset and the comparison results are shown in Fig. 7.

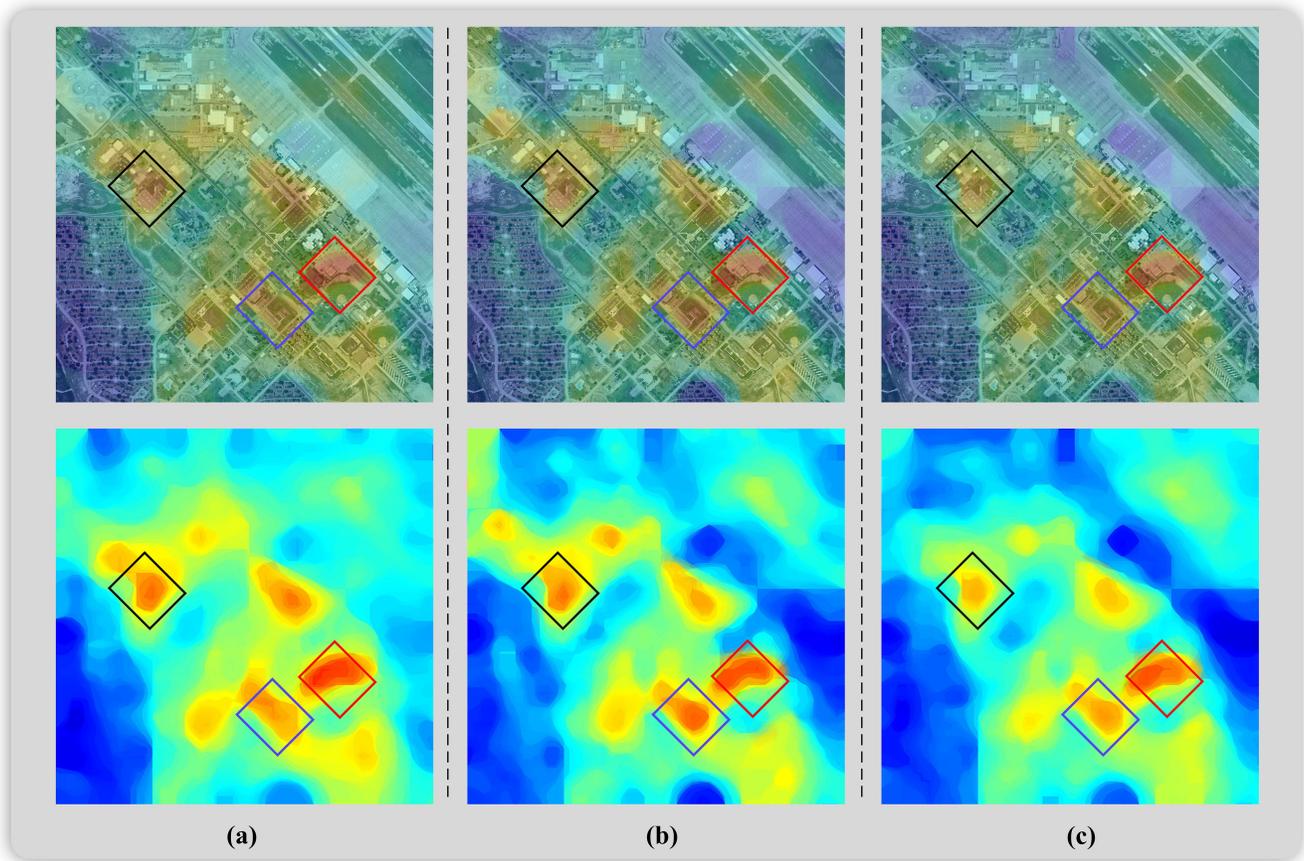

Fig. 8. Visualization comparison of semantic localization with query of "there are four red tennis courts, a green baseball field and a white parking lot surrounded by gray roads." (a) Only global information is used during inference. (b) The model adds global and local information to achieve semantic localization. (c) The MIDF module is leveraged to optimize the multi-level information. The area enclosed by the red box is the ground truth.

When perform image-text task, there tends to be more incorrect text for LW-MCR, which is consistent with the quantified performance indicators. AMFMN directly uses global features for retrieval, resulting in fewer complete retrieval instances among the candidates. Take the top2 text of AMFMN as an example, the model retrieves football but not baseball. GaLR model is based on the combination of global and local features, so that the information about instances in the queried RS image is often well-represented in the retrieval results. For example, the top3 text retrieved by the GaLR model has attributes such as playground and baseball, even though the text is not the ground truth.

The performance of the three models when using text to retrieve RS images also reflects the above conclusions. When retrieved by LW-MCR, although the ground truth is in the top5 images, some wrong categories are still retrieved. As a lightweight retrieval model, LW-MCR is more difficult to obtain better visual representation by feature extraction, which leads to false retrievals. Although AMFMN has retrieved the ground truth, most of the other images retrieved have not the full attributes of the queried text. Since LW-MCR can capture the local features of the RS image, it has a more fine-grained retrieval performance when conducting feature registration. The above experiments verify the desirability of GaLR in RS cross-modal retrieval tasks.

### H. Qualitative Analysis of Semantic Localization

In this subsection, to qualitatively demonstrate the performance of different configurations of GaLR on retrieval tasks, we conduct semantic localization experiments to visualize the impact of submodules on the results. The semantic localization task refers to the model locates the region that best matches the text in a large scene. Following Yuan *et al.* [14], after cutting the large scene image with a multi-scale sliding window, the probability distribution between the text and each slice is calculated. Then the obtained probability distributions are merged and the median filter is used to remove the impact noise in the probability map. Three different configurations of GaLR are utilized to accomplish this task.

- **(a)** Only global information is used during inference, which should be noted that the GaLR in this mode is equivalent to AMFMN.
- **(b)** The model adds GaLR to achieve semantic localization.

- **(c)** The MIDF module is leveraged to optimize the multi-level information.

The control group (a) and (b) shows the effect of adding local features on the retrieval performance, control group (b) and (c) demonstrates the performance improvement of the proposed MIDF module on the semantic localization task. While control group (a) and (c) demonstrates the direct comparison of the two methods of AMFMN and GaLR.

Fig. 8 shows the visualization results of the above three sets of experiments, where the ground-truth values are surrounded by red boxes. Each group is located image and heatmap from top to bottom, red represents the more likely to happen, blue is the opposite. In this experiment, the query is "There are four red tennis courts, a green baseball field and a white parking lot surrounded by gray roads," which is a relatively fine-grained sentence. We can draw the following conclusions.

- Compared with (a), the blue box area of (b) in the non-ground-truth region is significantly larger, which indicates that the model has judged more non-ground-truth regions after obtaining local representations. Compared with the black box area of (a), the attention of (b) in this area is weakened, and the model produces fewer false detection. The global representation of the model in these places may be more consistent with the query, but the similarity drops significantly when local features which reflect the details are taken into account. The false detection of the model begins to shift from the black box to the blue box area near the ground truth, indicating that the model has more confidence in the ground-truth area after obtaining local information.
- Compared to (b), (c) uses MIDF to dynamically select and fuse multi-level information. Compared with (b), the non-ground-truth area of (c) is enlarged again, indicating that the model pays more attention to the ground-truth area. Simultaneously, the attention of (c) on the black box and the blue box area is further reduced. MIDF dynamically filters the multi-level information when processing these regions, so that the model can calculate the similarity adaptively.
- (a) and (c) reflect the retrieval performance comparison between the proposed GaLR and AMFMN. Compared with (a), the attention of (c) on non-ground-truth regions is significantly reduced, which is sufficient to reflect the performance of the proposed GaLR in semantic localization and cross-modal retrieval tasks.

### I. Analysis of Time Consumption

In this subsection, we compare the retrieval time of the proposed model with other methods. The experiment platform is Intel(R) Xeon(R) Gold 6226R CPU @2.90 GHz and a single RTX 3090 GPU. To make a reasonable time comparison, two evaluation indicators are utilized: inference time (IT) and evaluation time (ET). IT refers to the time of performing a complete cross-modal similarity calculation. While ET refers to the time of calculating the similarity of all images and texts in different test sets. We conduct multiple experiments with no other load on the equipment to ensure fairness.

Table V shows the comparison of each model in terms of retrieval time. GaLR is slightly inferior to AMFMN in ET and at the same level as MTFN, which uses feature fusion to calculate similarity. Despite LW-MCR has great advantages in retrieval time, its performance is still far inferior to that of the GaLR. In terms of IT, the proposed method is more advanced than the MTFN but slightly less than AMFMN. On the whole, the retrieval time of GaLR is almost equal to that of other models, which belongs to the retrieval model of medium time consumption.

## V. Conclusion

In this article, an RS cross-modal retrieval framework based on GaLR is proposed. We optimize the representation matrix and the adjacency matrix of local features, and obtain SoTA retrieval results after combining multi-level features. In addition, a plug-and-play MR algorithm has been devised, which can improve the retrieval results without additional training. The quantitative analysis on multiple RS text-image datasets demonstrates the effectiveness of the proposed method for RS retrieval.

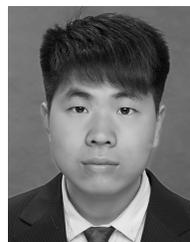

**Zhiqiang Yuan** (Student Member, IEEE) received the B.Sc. degree from Harbin Engineering University, Harbin, China, in 2019. He is pursuing the Ph.D. degree with the Aerospace Information Research Institute, Chinese Academy of Sciences, Beijing, China.

His research interests include multi-modal remote sensing image interpretation and multi-modal signal processing.

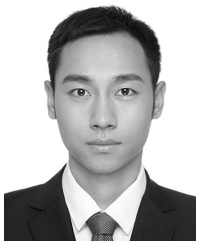

**Wenkai Zhang** (Member, IEEE) received the B.Sc. degree from the China University of Petroleum, Qingdao, China, in 2013, and the Ph.D. degree from the Institute of Electronics, Chinese Academy of Sciences, Beijing, China, in 2018.

He is an Assistant Professor with the Aerospace Information Research Institute, Chinese Academy of Sciences. His research interests include multi-modal signal processing, image segmentation, and pattern recognition.

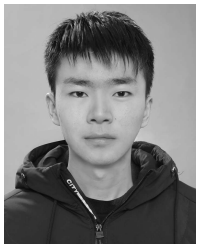

**Changyuan Tian** received the B.Sc. degree from Harbin Engineering University, Harbin, China, in 2021. He is pursuing the Ph.D. degree with the Aerospace Information Research Institute, Chinese Academy of Sciences, Beijing, China.

His research interests include remote sensing image understanding, deep learning, and computer vision.

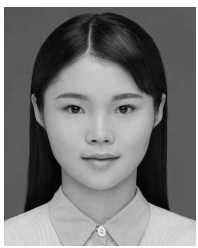

**Xuee Rong** received the B.Sc. degree from the Minzu University of China, Beijing, China, in 2019. She is pursuing the Ph.D. degree with the Aerospace Information Research Institute, Chinese Academy of Sciences, Beijing.

Her research interests include computer vision, pattern recognition, and remote sensing image processing, especially on continual semantic segmentation.

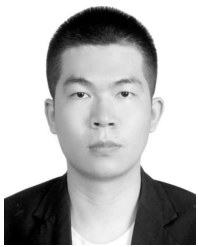

**Zhengyuan Zhang** received the B.Sc. degree from Harbin Engineering University, Harbin, China, in 2016. He is pursuing the Ph.D. degree with the Aerospace Information Research Institute, Chinese Academy of Sciences, Beijing, China.

His research interests include computer vision, pattern recognition, and remote sensing image processing, especially on image captioning.

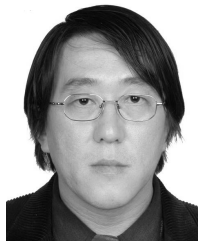

**Hongqi Wang** (Member, IEEE) received the B.Sc. degree from the Changchun University of Science and Technology, Changchun, China, in 1983, the M.Sc. degree from the Changchun Institute of Optics, Fine Mechanics and Physics, Chinese Academy of Sciences, Changchun, in 1988, and the Ph.D. degree from the Institute of Electronics, Chinese Academy of Sciences, Beijing, China, in 1994.

He is a Professor with the Aerospace Information Research Institute, Chinese Academy of Sciences. His research interests include computer vision and remote sensing image understanding.

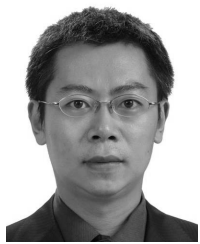

**Kun Fu** (Member, IEEE) received the B.Sc., M.Sc., and Ph.D. degrees from the National University of Defense Technology, Changsha, China, in 1995, 1999, and 2002, respectively.

He is a Professor with the Aerospace Information Research Institute, Chinese Academy of Sciences, Beijing, China. His research interests include computer vision, remote sensing image understanding, geospatial data mining, and visualization.

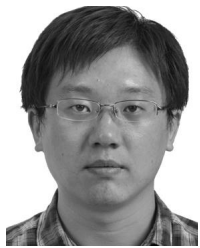

**Xian Sun** (Senior Member, IEEE) received the B.Sc. degree from the Beijing University of Aeronautics and Astronautics, Beijing, China, in 2004, and the M.Sc. and Ph.D. degrees from the Institute of Electronics, Chinese Academy of Sciences, Beijing, in 2009.

He is a Professor with the Aerospace Information Research Institute, Chinese Academy of Sciences. His research interests include computer vision, geospatial data mining, and remote sensing image understanding.